\documentclass[a4paper]{article}
\usepackage{Odyssey2024}
\usepackage{epsfig,amssymb,amsmath}
\usepackage{multirow}
\usepackage{booktabs} 
\usepackage{caption} 
\ninept

\title{Low-resource speech recognition and dialect identification of Irish in a multi-task framework
}

\name{$^1$Liam Lonergan, $^2$Mengjie Qian, $^1$Neasa Ní Chiaráin, $^1$Christer Gobl, $^1$Ailbhe Ní Chasaide}

\address{$^1$Phonetics and Speech Laboratory, School of Linguistics, Speech and Communication Sciences,\\ Trinity College Dublin  \\
$^2$Engineering Department, Cambridge University, UK  \\
{\small \tt $^1$\{llonerga, nichiarn, cegobl, anichsid\}@tcd.ie, $^2$mq227@cam.ac.uk}}
\begin{document}
\maketitle

\newpage
\begin{abstract}
This paper explores the use of Hybrid CTC/Attention encoder-decoder models trained with Intermediate CTC (InterCTC) for Irish (Gaelic) low-resource speech recognition (ASR) and dialect identification (DID). Results are compared to the current best performing models trained for ASR (TDNN-HMM) and DID (ECAPA-TDNN). An optimal InterCTC setting is initially established using a Conformer encoder. This setting is then used to train a model with an E-branchformer encoder and the performance of both architectures are compared. A multi-task fine-tuning approach is adopted for language model (LM) shallow fusion. The experiments yielded an improvement in DID accuracy of 10.8\% relative to a baseline ECAPA-TDNN, and WER performance approaching the TDNN-HMM model. This multi-task approach emerges as a promising strategy for Irish low-resource ASR and DID.
\end{abstract}

\section{Introduction}

Irish (Gaelic) is a highly inflected, minority language indigenous to the island of Ireland. A difficulty in developing speech technology for the language is the high degree of dialect variation across different speaker communities. There are three major native dialects of Irish, namely Ulster (Ul), Connaught (Co) and Munster (Mu), along with many sub-varieties. The L1 Irish speaking communities are found in limited remote areas called \textit{Gaeltachtaí}. However, there is  a considerable number of L2 speakers as Irish is also taught as a compulsory subject in primary and secondary schools. This paper focuses on the native dialects and the data used is overwhelmingly of L1 speakers.

Irish, being a low-resource language, poses a tough challenge for automatic speech recognition (ASR). Limited data makes it hard to create accurate and adaptable ASR models for Irish, impacting their effectiveness.  Additionally, the fact that there is no single standard spoken variety of Irish adds to the complexity: it is therefore crucial to identify these dialects to optimise the ASR process. As part of the ABAIR initiative, which is developing speech technologies for the Irish language, a TDNN-HMM ASR system \textit{Éist} \cite{lonergan2022automatic} has been developed and is available for public use\footnote{www.abair.ie}.  

Previous work established that Irish dialect bias in ASR could not be adequately mitigated through corpus balancing alone \cite{lonergan23_interspeech}. Spoken dialect identification (DID) of Irish was explored in \cite{lonergan2023sigul}, where an ECAPA-TDNN model achieved a classification accuracy of 73\%, surpassing the performance of wav2vec 2.0 XLS-R \cite{babu22_interspeech}  after finetuning. Further to this result, the output logits of a text-based classifier were fused with the acoustic classifier which improved classification accuracy to 76\%, demonstrating the importance of dialect features for classification in both the acoustic and the text domain. This paper aims to explore innovative approaches that could improve both ASR and DID performance in a multi-task setting. 

Multi-task training is a promising strategy that uses shared knowledge across related tasks. It has been shown to be an effective method for language identification in multilingual ASR systems~\cite{toshniwal2018improving}. Recognizing its potential to enhance model adaptability, this paper explores Intermediate CTC, which has been used for multilingual speech recognition \cite{chen2023improving}, for multi-task Irish speech recognition and dialect identification.

\section{Background}
Spoken language identification (LID) is the process of automatically identifying the language of a speaker from speech. Classical i-vectors \cite{martinez11b_interspeech, dehak2010front} were the state-of-the-art for LID before the introduction of acoustic embeddings extracted from DNNs, namely x-vectors \cite{snyder2018x}. Phonetically-aware acoustic features from acoustic models pretrained for ASR have been shown to outperform classical acoustic features for LID. Phonetic features extracted from a phone-discriminative model, initially trained as an ASR acoustic model, were explored in \cite{tang2017phonetic}, and led to an improvement over conventional acoustic features. Similarly, in \cite{rao2021icassp}, ASR-based phone posteriorgram features were used for accent identification of English, and improved performance over filter-bank features. In \cite{duroselle2021modeling} a Conformer model was initially trained for ASR with CTC, and subsequently used as a feature extractor, which yielded comparable and in some cases better acoustic features than classical multilingual bottleneck features for LID, without the need for phone-alignment information to train the feature extractor. The winners of the 2021 Oriental Language Recognition challenge~\cite{wang2021olr} pretrained an encoder-decoder U2++ Conformer model with ASR, before finetuning the encoder component for LID, demonstrating that conditioning an LID model to be phonetically-aware is a strong strategy~\cite{lyu22_interspeech}.

\cite{punjabi2021joint} explored jointly training an RNN-T model for language identification and ASR. It was found that the best jointly trained model in their study surpassed the performance of monolingual ASR by 6.4–9.2\% WER, and surpassed the LID system with a reduction in error rate of 53.9–56.1\%.

However, identifying the dialects within a single language is likely to be a more difficult task than the identification of completely separate languages. \cite{imaizumi2022end} investigates many configurations of combining ASR and DID, but most notably, their Japanese multi-dialect ASR system, which models ASR and DID in a multi-task set-up, outperforms the baseline system. Extending the vocabulary of a grapheme-based multi-dialect E2E ASR model for English to include a label for the dialect of a speaker is explored in \cite{li2018joint}, and this multi-task approach was found to outperform models trained on single dialects. 

Intermediate CTC (InterCTC) was originally proposed as a regularisation technique for deep encoders by incorporating the CTC loss of the intermediate layers as part of a multi-task objective~\cite{lee2021intermediate, sanabria2018hierarchical}. It has been proven to be effective for the joint modelling of ASR and a speech classification task, as in~\cite{chen2023improving} for multilingual speech recognition and language identification, achieving state-of-the-art results. Similarly, this method also achieved state-of-the-art Aphasia speech recognition and detection performance in \cite{tang23b_interspeech}.

In this paper, we explore the use of a hybrid CTC/Attention-based encoder-decoder model for Irish dialect identification and speech recognition. We explore the usefulness of InterCTC for both of these tasks by introducing an auxiliary task to predict dialect. The assignment of the auxiliary task to different encoder layers is varied systematically to find the optimum setting. Improvements to the encoder architecture are investigated, and shallow fusion with a transformer language model (LM) fine-tuned for multi-task DID and ASR shallow fusion is tested.



\section{Data}
\label{section:data}

\subsection{Acoustic data}
The data used in these experiments has either been recorded by the ABAIR project or drawn from external sources and are summarised in Table~\ref{tab:speech_corpora}. The recordings made in-house are divided into two corpora: MíleGlór and Synthesis. The MíleGlór (A thousand voices) data collection platform was created with speech recognition development in mind. Dialect-appropriate textual prompts were selected or crafted for recording. The dataset of 45.7h and 369 speakers includes crowdsourced recordings where participants recorded utterances using their own devices in varied recording environments as well as live recordings where the authors had control over the microphones used and the acoustic environment. The meta-data pertaining to the linguistc background of speakers is also collected, enabling selection of L1 vs L2 speakers. The Synthesis corpus of 25.6h is comprised of recordings of 5 L1 speakers used to create the ABAIR synthetic voices. 

The audiobooks used are from different sources: recordings made at home by two Mu Irish speakers reading the books \textit{Mo Sgéal Féin} and \textit{An tOileánach} were used. A collection of stories published by \textit{Cois Life} as well as a collection of short stories from \textit{Éabhlóid} were used as well. A spontaneous speech corpus of broadcast material named \textit{Corpas na Cainte Beo} provided by Foras na Gaeilge’s New English-Irish Dictionary project, which is tagged with dialect information, is also used. This corpus was provided to ABAIR without audio-text alignment. The alignment procedure employed is detailed in Section~\ref{sec:alignment}.

\begin{table}[!htbp]
\caption{Speech corpora}
\label{tab:speech_corpora}
\centering
\begin{tabular}{l|c}
\toprule
Type    & Duration (h) \\
\midrule
Audiobooks           & 33.6  \\
Synthesis            & 25.6  \\
MíleGlór             & 45.7  \\
Corpas na Cainte Beo & 200.8 \\
\bottomrule
\end{tabular}
\end{table}

\subsubsection{Alignment of Spontaneous Speech Corpus}
\label{sec:alignment}
The alignment of the spontaneous speech corpus was done in two stages, firstly using CTC-Segmentation as described in~\cite{kurzinger2020ctc} using the ESPnet toolkit~\cite{watanabe2018espnet}. CTC-Segmentation utilizes CTC log-posteriors to determine utterance timings in the audio given a ground-truth text. Initially, a forward pass is executed, wherein transition probabilities are mapped into a trellis diagram of the ground-truth token sequence across all time steps. The algorithm backtracks from the most probable timing of the last token and finds the most probable path through the trellis diagram. A confidence score is computed for each utterance, based on per-token probabilities within the trellis. A Conformer encoder \cite{gulati20_interspeech} which uses XLS-R 300M as the frontend with the CTC objective is trained on all available Irish data and used for alignment. CTC-segmentation requires inference over the entire audio sequence of a file, which when using Transformer-based encoders with quadratic memory-based complexity, can be an issue for longer files. Following the implementation in~\cite{takamichi2021jtubespeech}, files above 600s are partitioned into smaller segments of audio. Inference is calculated on these segments and the CTC-posteriors of the segments are subsequently concatenated. As splitting audio abruptly can lead to distortions, an overlap of 1s is used and these overlap posteriors are later discarded for scoring.

The resulting alignment contained speech segments longer than the conventional maximum of 20s for ASR. To split these files into segments 20s in length, Kaldi-based scripts are used. A biased LM is created according to the transcripts of the input data and together with a TDNN-HMM acoustic model, the data is split into reasonable chunks\footnote{kaldi/egs/wsj/s5/steps/cleanup/segment\_long\_utterances\_nnet3.sh}. An additional step was taken to ensure the resulting splits match their transcripts by removing bad portions or to make minor modifications to allow for disfluencies or repetitions\footnote{kaldi/egs/wsj/s5/steps/cleanup/clean\_and\_segment\_data\_nnet3.sh}. Implementation details for both of these scripts can be found in~\cite{manohar2017jhu}. 200h out of a total 320h were successfully aligned using this process.

\subsubsection{Construction of train, validation and test sets}
The train set contains 290h in total, and the validation and test sets contain 1.7h and 3.5h respectively. These were constructed such that there was no overlap of speakers or utterance texts between the sets. As this paper is focusing on ASR and dialect identification of L1 speakers of Irish, the MíleGlór corpus is the most appropriate set to use to construct the validation and test sets. As noted, specific meta-data relating to the linguistic background of each speaker was  collected and therefore, including only L1 speakers of Irish in the test sets can be done easily. Also, the authors had control over the text prompts used when recording and could ensure their dialect appropriateness. One issue with this corpus which pertains to Ulster speakers particularly, is that only a limited set of 3000 recording prompts were used for Ul recordings. Therefore, there is less variability of utterance texts in the Ul portion of this corpus, compared to Co and Mu. Avoiding overlap of texts would necessitate a large portion of Ul data to be discarded from any set. To avoid this, a portion 0.2h of Ul data from the Audiobook collection was added to the test set. Details of the validation and test sets are presented in Table~\ref{tab:text_corpora}.

\begin{table}[!htbp]
\caption{Breakdown of validation and test sets with number of speakers and duration}
\label{tab:text_corpora}
\centering
\begin{tabular}{l|cc|cc}
\toprule
        & \multicolumn{2}{c|}{Validation set} & \multicolumn{2}{c}{Test set}             \\
Dialect & \#spks & dur (h) & \#spks & dur (h) \\
\midrule
Ul      & 9      & 0.55h         & 17     & 1.03h         \\
Co      & 4      & 0.55h         & 12     & 1.29h         \\
Mu      & 5      & 0.55h         & 20     & 1.15h        \\
\midrule
Total   & 18    & 1.66h          & 49     & 3.47h \\
\bottomrule
\end{tabular}
\end{table}
\subsection{Text corpora for shallow-fusion experiment}
\label{section:lm}
The transformer language model used in the shallow-fusion experiment is trained in two stages. Firstly, the model is trained with text-only data, and then it is fine-tuned with corpora containing dialect information. In the first stage, two corpora were used: i) Paracrawl, which is the Irish part of the ga-en pairs from ParaCrawl v7 \cite{banon-etal-2020-paracrawl}; and ii) ConLL17, the Irish data from the CoNLL 2017 Shared Task on Universal Dependancy Parsing \cite{zeman-etal-2017-conll}. As these corpora were scraped from the web, they contain many symbols and characters from different languages. There is only a limited amount of textual data available on the web in the Irish language and therefore, the two corpora contain some overlap. To overcome these issues, an aggressive multi-stage cleaning process was employed: i) a conventional text-cleaning script for Irish was used, which transliterates certain characters and removes unnecessary punctuation symbols. After this, any sentence that contained symbols outside of the limited character set of Irish was removed. The corpus was then sorted and uniqued on a sentence level to ensure that overlap between the two corpora is resolved. This resulted in a corpus of 58m words, a reduction of almost 50\% of the combined size of the Paracrawl and CoNLL 2017 and is used to pretrain the transformer LM used for shallow fusion.

The second collection of texts that were used include dialect information at the sentence level, which is generally not available in Irish corpora. One useful source that is tagged with dialect is the Historical Irish Corpus of the Royal Irish Academy. This corpus consists of texts written between 1900-50, before the introduction of the standard written form, so morphosyntactic and lexical markers of dialect are salient in these texts (for more details, see \cite{lonergan2023sigul}). Alongside this corpus, the transcripts from the entire Spontaneous Speech Corpus were used, as well as the transcripts from the training set. See for Table~\ref{tab:text_lm} for information regarding these corpora.

\begin{table}[!htbp]
\caption{Text corpora used for LM training and multi-task fine-tuning}
\centering
\label{tab:text_lm}
\begin{tabular}{l|ll}
\toprule
            & Corpus     & \#words \\
\midrule
\multirow{2}{*}{Training}    & Paracrawl  & 63.1m   \\
            & ConLL17    & 49.4m   \\
\midrule
\multirow{2}{*}{Fine-tuning} & Historical & 4.5m    \\
            & Spontaneous Speech  & 4.4m    \\
            & Training   & 0.9m   \\
\bottomrule
\end{tabular}
\end{table}

\section{Method}
\subsection{Hybrid CTC/Attention}
Hybrid \text{CTC}/\text{Attention}-based encoder-decoder models \cite{watanabe2017hybrid} utilize two principal techniques in ASR, namely Connectionist Temporal Classification CTC \cite{graves2006connectionist} and the attention mechanism. The process can be summarized as follows:

The encoder, denoted as a function \(E\), transforms an acoustic sequence \(X = \{x_1, x_2, \ldots, x_T\}\) into a series of embeddings \(H = \{h_1, h_2, \ldots, h_N\}\), where \(T\) represents the length of the input sequence, and \(N\) is the length of the encoded embeddings.
\begin{equation}
\bold{H} = E(X)
\end{equation}
These embeddings \(H\) can optionally be converted into text using CTC, generating a preliminary textual output \( P_{\text{Enc}}(T | X) \).
\begin{equation}
P_{\text{Enc}}(T \,|\, X) = {\text{CTC}(\bold{H})}
\end{equation}

The decoder, utilizing an auto-regressive approach and denoted by \(Dec\), predicts the text output \(Y = \{y_1, y_2, \ldots, y_K\}\) given the acoustic embeddings \(H\). 
\begin{equation}
P(y_k | X, Y_{1:k-1}) = \text{Dec}(\bold{H}, T_{1:k-1})
\end{equation}
\begin{equation}
P_{\text{Dec}}(T | X) \approx \prod_{k}^KP(t_k | X, T_{1:k-1})
\end{equation}

During training, the model is optimized using the weighted sum of the CTC loss and the decoder loss, with a weight set to 0.3 . The final hypothesis is generated during inference by jointly decoding the output of the encoder and decoder using beam search. This hybrid approach effectively combines the robust alignment capabilities of CTC with the contextual sensitivity of the attention mechanism to improve ASR performance.

\subsection{Intermediate CTC}
\label{sec:inter_ctc}

Intermediate CTC (InterCTC) \cite{lee2021intermediate} was introduced as a regularization technique for deep encoder networks and to facilitate multi-task learning \cite{chen2023improving,tang23b_interspeech}. A CTC module is applied to the output of an intermediate encoder layer with index $e$. Self-conditioned CTC is also applied, where subsequent encoder layers incorporate these intermediate predictions into their input. The rearranged Equation 1 is expressed as follows:
\begin{equation}
    \bold{H}_e = \text{Enc}_{1:e}(\bold{X}) \label{eq:he} 
\end{equation}
\begin{equation}
    P(Z_{\text{Inter}}|X) = \text{CTC}(\bold{H}_e) \label{eq:pzinter}
\end{equation}
\begin{equation}
    \bold{H} = \text{Enc}_{e+1:E}(\texttt{NRM}(\bold{H}_e) + \texttt{LIN}(Z_{\text{Inter}})) \label{eq:h}
\end{equation}
where
$E$ represents the encoder layers, and $Z_{\text{Inter}}$ is the latent sequence of the InterCTC target sequence $T_{\text{Inter}} = (t'_k|k = 1, ..., K')$. The functions \texttt{NRM(·)} and \texttt{LIN(·)} correspond to a normalization layer and a linear layer, respectively. The InterCTC loss is the negative log likelihood of generating $T_{\text{Inter}}$:
\begin{equation}
    \mathcal{L}_{\text{Inter}} = - \log P_{\text{Inter}}(T_{\text{Inter}}|X) \label{eq:linter}
\end{equation}
The selection of $T_{\text{Inter}}$ depends on the task. During training, the intermediate layer is optimized to accurately predict $T_{\text{Inter}}$ by integrating $\mathcal{L}_{\text{Inter}}$ into the loss function:
\begin{equation}
    \mathcal{L}'_{\text{CTC}} = \alpha L_{\text{Inter}} + (1 - \alpha) \mathcal{L}_{\text{CTC}} \label{eq:lctc}
\end{equation}
Here, the InterCTC weight $\alpha$ serves as a hyper-parameter. The updated overall loss function is derived by inserting Equation \ref{eq:lctc} into Equation 5:
\begin{equation}
    \mathcal{L}' = \lambda \mathcal{L}'_{\text{CTC}} + (1 - \lambda) \mathcal{L}_{\text{Dec}} \label{eq:lprime}
\end{equation}
It is noteworthy that CTC can be applied to multiple encoder layers with different target sequences for each. Where the target sequence differs from the decoder output, this is considered as an auxiliary task. The average of all InterCTC losses is used as $\mathcal{L}_{\text{Inter}}$.

\subsection{Dialect identification as an auxiliary task}
Using the above model structure, dialect is captured by a combination of two methods. Firstly, it is captured explicitly in the text output sequence, by prepending a dialect tag to the text to be predicted:
\begin{equation*}
\text{[CO] \textit{anois teacht an earraigh}}    
\end{equation*}
The vocabulary of the ASR model is extended to include dialect tags in the following way:
\begin{equation}
    V' = V \cup \{ [UL], [CO], [MU] \}
\end{equation}
This approach jointly models the ASR task and the speech classification task effectively. 

The second method to capture dialect is to use a DID InterCTC objective, where the ground truth of an utterance is the dialect tag of the speaker. During inference, the dialect of an utterance is predicted using InterCTC greedy search.

\subsection{Model structure}
The model training in these experiments adopts the hybrid CTC/Attention-based encoder-decoder framework. Two encoder architectures, namely the Conformer and the E-branchformer~\cite{kim2023branchformer}, are investigated. These encoders were proposed as improvements to the transformer encoder architecture, which captures global acoustic contextual information using the attention mechanism. The Conformer in addition to utilizing convolutional layers to capture local contexts, which is particularly important for speech, also makes use of the attention mechanism to capture global information. On the other hand, the E-branchformer has superseded the Conformer encoder as the state-of-the-art encoder, by capturing the local and global contexts in parallel branches before merging the outputs. For the front-end module, a self-supervised learning model (SSL) was chosen, namely the multilingual XLS-R 300M~\cite{babu22_interspeech}, which is trained on 436k hours of data from 128 languages. Despite the absence of Irish in this multilingual training set, this model was deemed to be more appropriate for Irish speech recognition and the dialect disambiguation than other acoustic SSL models typically trained monolingually on English data. All models trained use a transformer decoder.

\section{Experiments}
\subsection{Setup}
\label{section:setup}
In these experiments, the hybrid CTC/Attention-based encoder-decoder structure is explored for jointly modelling Irish speech recognition and dialect identification, using the ESPnet toolkit~\cite{watanabe2018espnet}. See Section 4.3 for details on how dialect is incorporated. Accuracy is used to report DID performance, as the test set is quite balanced with respect to the number of utterances per dialect. Throughout all experiments, the XLS-R 300M was used as a front-end module for acoustic feature extraction. All encoder-decoder models are trained with a transformer decoder with 6 blocks, each having 2048 hidden units and 4 attention heads. Three different encoders are tested in these experiments, a Conformer encoder and a small and large E-branchformer encoder. The Conformer encoder has 12 blocks, each having 2048 hidden units and 4 attention heads, and Conformer-based models trained here have 113m trainable parameters. The small E-branchformer model has 45.8m trainable parameters, with 12 blocks, each with 1048 linear units, an output size of 256 and 4 attention heads. The larger model has 130m trainable parameters and uses 2048 linear units and an output size of 256, while the remaining parameters are kept the same. The cgMLP module of the E-branchformer encoders has 3072 units and the convolution kernel size is 31. Speed perturbation with warping factors of 0.9, 1.0 and 1.1 as well as Spectral Augment are used to augment the training data.

A modular TDNN-HMM ASR model is trained with speed perturbation and spectral augment  as a comparison for ASR performance and an ECAPA-TDNN model is trained with speed perturbation and added noise and reverberation as a comparison for DID performance.

\subsection{ASR baseline models}
Despite the advance of End-to-End ASR models, the hybrid TDNN-HMM is still a robust model for low-resource languages and was still the best performing model for Irish~\cite{lonergan22_interspeech}. A TDNN-HMM model was trained as the baseline to compare with the CTC/Attention encoder-decoder architecture, following  the same set-up as \cite{lonergan22_interspeech}. A 4-gram LM is trained using both the training and fine-tuning text corpora listed in Section \ref{section:lm} and Table~\ref{tab:text_lm}.

To evaluate both the effectiveness of InterCTC on DID and ASR performance and the impact of multi-task learning on ASR performance, three Conformer-based models are trained and compared with the hybrid TDNN-HMM baseline. The first Conformer-based model, which will be referred to as Conformer (ASR) and is the second row in Table~\ref{tab:asr_baseline}, is trained for ASR only with an ASR InterCTC objective in layers 3, 6 and 9. This model is designed to compare the capability of the model architecture on the ASR task with TDNN-HMM on the ASR task. The second Conformer model, which will be referred to as Conformer (multi-task) and is  the third row in Table~\ref{tab:asr_baseline}, is trained for multi-task ASR and DID with InterCTC DID and ASR objectives defined for encoder layers 3, 6 and 9. 
The third model Conformer (no InterCTC), see the fourth row in Table~\ref{tab:asr_baseline}, is trained for multi-task ASR and DID without InterCTC.

Table~\ref{tab:asr_baseline} presents the ASR performance of these models. The TDNN-HMM baseline outperforms the best performing Conformer model, specifically Conformer (ASR), by 16.5\% relative WER. Conformer (multi-task) performs worse than Conformer (ASR) by 13\% relative, suggesting that jointly modelling DID and ASR leads to ASR performance degradation. Conformer (no InterCTC) performs less well, showing that the addition of InterCTC leads to WER improvements.

\begin{table}[!htbp]
\caption{Performance of ASR baselines.}
    \label{tab:asr_baseline}
    \centering
    \begin{tabular}{lll||c}
    \toprule
    Model   & Objective & InterCTC &  {WER} \\
    \midrule
    TDNN-HMM & & & 13.2  \\
    Conformer & ASR & yes & 15.8  \\
    Conformer & ASR + DID & yes  & 18.1 \\
    Conformer & ASR + DID & no & 18.9  \\
    \bottomrule
    \end{tabular}
\end{table}

\subsection{DID baseline models}
Previous work \cite{lonergan2023sigul} demonstrates that the ECAPA-TDNN performs well for Irish DID, outperforming wav2vec 2.0, hence an ECAPA-TDNN model was trained using SpeechBrain \cite{speechbrain} following the same setup as \cite{lonergan2023sigul}, serving as the baseline model for the DID task. The embedding model is initialised from an ECAPA-TDNN trained for language identification using the VoxLingua107 corpus \cite{valk2021slt}.  The model is trained using the Additive Angular Margin loss \cite{deng2019arcface} and speed perturbation, adding noise and reverberation are applied as data augmentation.

In Table~\ref{tab:baseline_did}, the DID performance of the ECAPA-TDNN is compared with a Conformer (multi-task) and a Conformer (no InterCTC). Both Conformer models outperform the ECAPA-TDNN model by a wide margin. Interestingly, InterCTC with the multi-task objective assigned to layers 3, 6 and 9 does not lead to a gain in DID accuracy compared to the model trained without InterCTC.



\begin{table}[!htbp]
\caption{Performance of DID baselines.}
    \label{tab:baseline_did}
    \centering
    \begin{tabular}{l||cccc}
    \toprule
    Model      & DID Acc. \\
    \midrule
    ECAPA-TDNN & 72.5 \\ 
    Conformer (multitask)  & 78.6  \\
    Conformer (no InterCTC) & 79.7 \\
    \bottomrule
    \end{tabular}
\end{table}

\subsection{Experiment 1}
In this experiment, assigning InterCTC objectives to layers of the Conformer encoder is explored.  The assignment of the Multi-task and DID InterCTC objectives, detailed in Section 4.3, to the encoder layers is varied systematically. An additional model trained to perform both ASR and DID without InterCTC is included here for comparison.

Table~\ref{tab:InterCTC_layer} shows the results of the InterCTC configurations for DID accuracy and WER. Row 1 shows the performance of baseline Conformer (no InterCTC), and row 7 shows the performance for baseline Conformer (multi-task). As mentioned before, the Conformer (no InterCTC) model  achieved a DID accuracy of 79.7\%. However, in comparison with the models trained with InterCTC, its ASR performance is relatively poor at 18.9\% WER, suggesting that InterCTC is most helpful for ASR performance as opposed to the speech classification task of DID. Surprisingly, the models trained with DID InterCTC objectives only, as in rows 2-4, performed worse in DID accuracy than the baseline on row 7, which was trained with multi-task CTC objectives. However, the ASR performance of the these models (rows 2-4) varies considerably: row 2 trained with DID InterCTC objective in layers 3, 6 and 9 garnered the best WER (16.6\%) in this experiment;  row 4, where the only InterCTC objective was assigned to layer 3, achieved the worst WER in the experiment (19.4\%). The best performing system (row 6) is trained with the DID objective in layer 3 and the multi-task objective in layers 6 and 9. This model achieved the highest DID accuracy in this experiment, a boost of 10.8\% relative to the ECAPA-TDNN baseline. The same model also achieved the second lowest WER among the models trained for both ASR and DID, with 16.7\% WER. This model is chosen as the best configuration of InterCTC objectives for joint DID and ASR modelling for Experiments 2 and 3.

\begin{table}[!htbp]
\caption{DID and ASR performance with varying InterCTC settings. Row 1 is Conformer (no InterCTC) and row 7 is Conformer (multi-task) from the baseline experiments}
    \label{tab:InterCTC_layer}
    \centering
    \begin{tabular}{@{ }c|l@{ }@{ }l@{ }|c|cccc@{}}
    \toprule
    & \multicolumn{2}{c|}{Inter-CTC} & DID & \multicolumn{4}{c}{WER} \\
    &     Multitask  & DID  & Acc & All & Ul & Co & Mu \\
    \midrule
    1* & \multicolumn{2}{c|}{-} & 79.7 & 18.9 & 18.6 & 20.7 & 17.2 \\
    2 & - & L 3, 6, 9 & 78.4 & 16.6 & 16.2 & 18.0 & 15.4 \\
    3 & - & L 3, 6 & 69.1 & 18.7 & 17.9 & 20.1 & 17.8 \\
    4 & - & L 3 & 75.3 & 19.4 & 19.3 & 20.0 & 18.4 \\
    5 & L 9 & L 3, 6 & 77.0 & 18.2 & 17.8 & 19.3 & 17.3 \\
    6 & L 6, 9 & L 3 & 80.3 & 16.7 & 16.4 & 18.2 & 15.2 \\
    7* & L 3, 6, 9 & -  & 78.6 & 18.1 & 17.8 & 19.3 & 17.1 \\
    \bottomrule
    \end{tabular}
\end{table}

\subsection{Experiment 2}
The E-branchformer architecture using the best InterCTC configuration established in Experiment 1 is explored to compare the performance of the Conformer and the E-branchformer encoder. Two models are trained using the E-branchformer encoder - a smaller version (45.8m trainable parameters) and a larger version (130m trainable parameters). For further details see Section~\ref{section:setup}.

The best configuration (row 6 in Table~\ref{tab:InterCTC_layer}) of Experiment~1 is selected to compare the performance of Conformer and E-branchformer encoders in this experiment. Results are presented in Table~\ref{tab:ebranch}. E-branchformer Small is the best performing model in terms of DID, with an accuracy of 81.4\%, an improvement of 1.1\% absolute. The model's WER performance however is worse, suggesting that the smaller number of trainable parameters more strongly affects WER performance than it does DID performance. E-branchformer Large performs slightly worse than the Conformer model in terms of DID, but it has a relative WER performance gain of 6\%.

\begin{table}[!htbp]
\caption{DID \& ASR performance for Conformer and E-branchformer encoders and multi-task LM shallow fusion.}
    \label{tab:ebranch}
    \centering
    \begin{tabular}{@{}l@{ }|c|cccc@{ }}
    \toprule
    Encoder     & DID & \multicolumn{4}{c}{WER} \\
         & Acc & All & Ul & Co & Mu \\
    \midrule
    \multirow{1}{*}{Conformer} & 80.3 & 16.7 & 16.4 & 18.2 & 15.2 \\
        \midrule
     E-Branchformer Small  & \textbf{81.5} & 17.7 & 17.6 & 19.1 & 16.3 \\
          E-Branchformer Large  & 80.0 & \textbf{15.7} & \textbf{15.5} & \textbf{17.0} & \textbf{14.5}\\
      \hspace{23mm} + LM  & 80.8 & \textbf{13.5} & \textbf{13.0} & \textbf{14.9} & \textbf{12.3} \\
    \bottomrule
    \end{tabular}
\end{table}

\subsection{Experiment 3}
This experiment explores the usefulness of multi-task shallow fusion. A transformer language model was initially trained with a larger corpus of Irish text and fine-tuned on a smaller dialect-tagged corpus of Irish text for multi-task DID and ASR shallow fusion. For details of the corpora used, see Table~\ref{tab:text_lm}. A grid-search found 0.3 to be the optimal weight for shallow fusion.

The results for LM shallow fusion can also be seen in Table~\ref{tab:ebranch}. A slight improvement in DID classification accuracy of 0.8\% was gained through shallow fusion, but its most significant contribution is, unsurprisingly, a reduction of WER by 2.2\%.

\section{Discussion and Conclusions}
The hybrid CTC/Attention-based encoder-decoder architecture emerges as a good strategy for Irish multi-task ASR and DID. The best results were obtained from the E-branchformer Large model with the optimal InterCTC setting (established in Experiment 1) and multi-task LM shallow fusion, where the LM was fine-tuned with dialect labels prepended to the text. That the inclusion of LM shallow fusion improves DID accuracy slightly is an encouraging finding. Perhaps with more available dialect-tagged text corpora, such an approach could garner further improvement. 

The present DID results are promising in that they surpass the previous best model trained (ECAPA-TDNN). The best DID accuracy obtained here is 81.5\% compared to 73\% obtained in~\cite{lonergan2023sigul}. The overall contribution of InterCTC to DID accuracy is slight, as can be seen by comparing row 1 in Table~\ref{tab:InterCTC_layer} with the other rows. Only in one case (row 6) does the inclusion of InterCTC yield a DID improvement. This differs from what was found in \cite{chen2023improving} and \cite{tang23b_interspeech}, where InterCTC improved the speech classification accuracy more considerably.

As regards WER, training for multi-task DID and ASR does lead to some performance degradation. However, when the InterCTC objectives are optimally set, the WER degradation is greatly lessened. While the ASR performance of these E2E models is not better than the modular TDNN-HMM model, the gap between the two approaches is reduced a great deal and this approach promises to be a fruitful avenue for continued exploration.

\section{Acknowledgements}
This work is part of the ABAIR initiative, which is supported by the Department of Tourism, Culture, Arts, the Gaeltacht, Sport and Media, with funding from the National Lottery, as part of the 20-year Strategy for Irish. The authors gratefully acknowledge the speakers and publishers that have contributed to this work.

\bibliographystyle{IEEEbib}
\bibliography{Odyssey2024}

%

\end{document}